\documentclass[sigconf]{acmart}
\usepackage{graphicx}
\usepackage{subfigure}
\usepackage{array}
\usepackage{hyperref}

\setcopyright{rightsretained}
\usepackage{todonotes}
\usepackage[compact]{titlesec}

\copyrightyear{2020} 
\acmYear{2020} 
\acmConference[CoDS COMAD 2020]{7th ACM IKDD CoDS and 25th COMAD}{January 5--7, 2020}{Hyderabad, India}
\acmBooktitle{7th ACM IKDD CoDS and 25th COMAD (CoDS COMAD 2020), January 5--7, 2020, Hyderabad, India}
\acmDOI{10.1145/3371158.3371403}
\acmISBN{978-1-4503-7738-6/20/01}

\begin{document}

\title{Memeify: A Large-Scale Meme Generation System}


\author{Suryatej Reddy Vyalla}
\authornote{Both authors contributed equally to this research.}
\affiliation{%
  \institution{IIIT Delhi, India}}
  \email{suryatej16102@iiitd.ac.in}

\author{Vishaal Udandarao}
\authornotemark[1]
\affiliation{%
  \institution{IIIT Delhi, India}}
\email{vishaal16119@iiitd.ac.in}

\author{Tanmoy Chakraborty}
\affiliation{%
  \institution{IIIT Delhi, India}}
  \email{tanmoy@iiitd.ac.in}

\renewcommand{\shortauthors}{Vyalla and Udandarao, et al.}

\begin{abstract}
Interest in the research areas related to meme propagation and generation has been increasing rapidly in the last couple of years.
Meme datasets available online are either specific to a context or contain no class information. Here, we prepare a large-scale dataset of memes with captions and class labels.
The dataset consists of $1.1$ million meme captions from $128$ classes. We also provide a  reasoning for the existence of broad categories, called `themes' across the meme dataset; each theme consists of multiple meme classes.
Our generation system uses a trained state-of-the-art transformer based model for caption generation by employing an encoder-decoder architecture. We develop a web interface, called {\bf Memeify} for users to generate memes of their choice, and explain in detail, the working of individual components of the system. We also perform qualitative evaluation of the generated memes by conducting a  user study. A link to the demonstration of the Memeify system is \href{https://youtu.be/P_Tfs0X-czs}{https://youtu.be/P\_Tfs0X-czs}.
\end{abstract}

    \maketitle

\section{Introduction}
Memes \cite{r.dawkins1976the-selfish-gen} are currently the hottest trending ways of expressing ideas and opinions on social media. Ever since the social media boom happened in the late 2000s, memes have conquered the Internet landscape in ways no one could have imagined. An average social media user has access to hundreds of memes every day on various platforms. Creating a meme requires context and creativity \cite{gatt2018survey}. Recent advances in deep learning and natural language processing have allowed neural networks to be used for generative tasks \cite{lehman2016creative} \cite{colombo2017deep} \cite{goodfellow2014generative}. 

 Peirson et al. \cite{peirson2018dank} were one of the first to propose a meme generation network by modeling the task of meme generation as an image captioning problem. Their dataset does not contain class labels. They used standard glove embeddings to create word vectors of captions and modelled the generation task using a simple encoder decoder architecture. 

There have been significant advances in deep natural language processing tasks ever since \cite{DBLP:journals/corr/abs-1907-11692} \cite{DBLP:journals/corr/abs-1906-08237}. The BERT \cite{devlin2018bert} model generates vectors for each word by training a transformer network to extract deep bidirectional representations. The MT-DNN \cite{liu2019multi} model generalizes the BERT bidirectional language model by applying an effective regularization mechanism and creates the word vectors. Our model leverages these latest techniques for generating memes. 

In this paper, we explain our proposed meme generation system, called {\bf Memeify} which can creatively generate captions given a context.
We first build a large-scale dataset that is both reliable and robust. Further, we draw some inferences on the characteristics of memes based on their image backgrounds and captions. Finally, we propose an end-to-end meme generation system that allows users to generate memes on our Memeify web application.

    The major contributions of the paper are two-fold: (i) generation of a large-scale meme dataset, which to our knowledge, is the first of its kind, and (ii) design of a novel web application to generate memes in real time. 
    
     For the sake of reproducible research, we have made the code and dataset public at \href{https://github.com/suryatejreddy/Memeify}{https://github.com/suryatejreddy/Memeify}.
\section{Dataset Description}

\begin{figure*}[!t]
  \centering
  \subfigure[TSNE plot of word clusters]{\includegraphics[width=0.35\linewidth,height=0.25\linewidth]{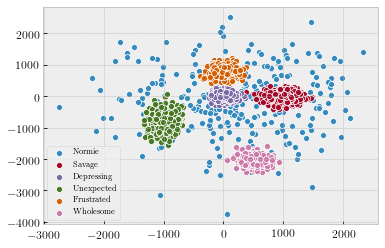}}\quad 
  \subfigure[TSNE plot of image clusters]{\includegraphics[width=.35\linewidth,height=0.25\linewidth]{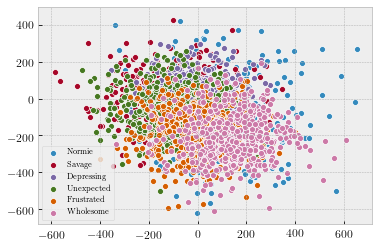}}
  \caption{Depiction of the meme clusters formed when applying image vectorization and word vectorization.}
  \label{fig:cluster}
\end{figure*}

\label{dataset}

\subsection{Data Curation}
We created our own dataset of meme images, their captions and corresponding class information. Every meme belongs to a particular class depending on the base image of the meme. The current datasets on memes available online have various limitations. The Reddit meme dataset on kaggle \cite{redditmemes} does not have captions for the memes and the class information.  The meme generator dataset \cite{memegendata} contains 90k images with captions and class labels. But there is a huge class imbalance and we could not download most of the images because the links were broken. 

To overcome these issues, we scraped data from QuickMeme \cite{quickmemeweb} and a few other sources to create a dataset of $1.1$ million memes belonging to $128$ classes. We have base images for all the classes and captions for each meme. We require such a large dataset to train our meme 
generation model.

\subsection{Dataset Analysis: Themes}
\label{theme_data}
To understand our data, we sample (stratified) 5,000 memes and for each meme, we create an average word embedding of the captions to perform clustering \cite{Berkhin2006ASO}. Visualization of these clusters is shown in Figure \ref{fig:cluster}(a). We observe that there are 5 distinct clusters in the data, and the remaining embeddings are equally spread out in the space. The results were consistent when we tried on different samples.

From this, we hypothesize that every meme can be associated with a broad category, which we call `theme', based on the words in the caption. We segregate meme classes into themes as identified by the clustering algorithm and labeled them as ``Savage", ``Depressing", ``Unexpected", ``Frustrated" and ``Wholesome". The remaining classes are labeled as ``Normie". This labeling is done on the basis of captions and their usage on social media platforms. A class $X$ is labelled theme $Y$ if more than 90\% of the memes from class $X$ belong to cluster $Y$. Using this condition, every class is assigned a cluster (theme) as shown in Table \ref{tab:themecluster}.

\begin{table}
\caption{Assignment of classes to themes.}
\label{tab:themecluster}
\vspace{-3mm}
\begin{tabular}{|l|l|}
\hline
\textbf{Theme} & \textbf{Count} \\ \hline
Normie         & 44             \\ \hline
Savage         & 22             \\ \hline
Depressing     & 18             \\ \hline
Unexpected     & 20             \\ \hline
Frustrated     & 14             \\ \hline
Wholesome      & 10             \\ \hline
\end{tabular}
\vspace{-5mm}
\end{table}

To confirm our hypothesis, we create image vectors for each of our memes using a VGG16 convnet \cite{simonyan2014very}. We then cluster these vectors and assign themes as labels for plotting. The result is shown in Figure \ref{fig:cluster}(b). We infer that the theme of a meme depends on the words in its caption and not the background image.  

\section{MEMEIFY: Our Proposed Architecture}

\begin{figure*}[!htp]
  \centering
  \subfigure[Memeify web application landing page]{\includegraphics[width=.40\linewidth,height=.30\linewidth]{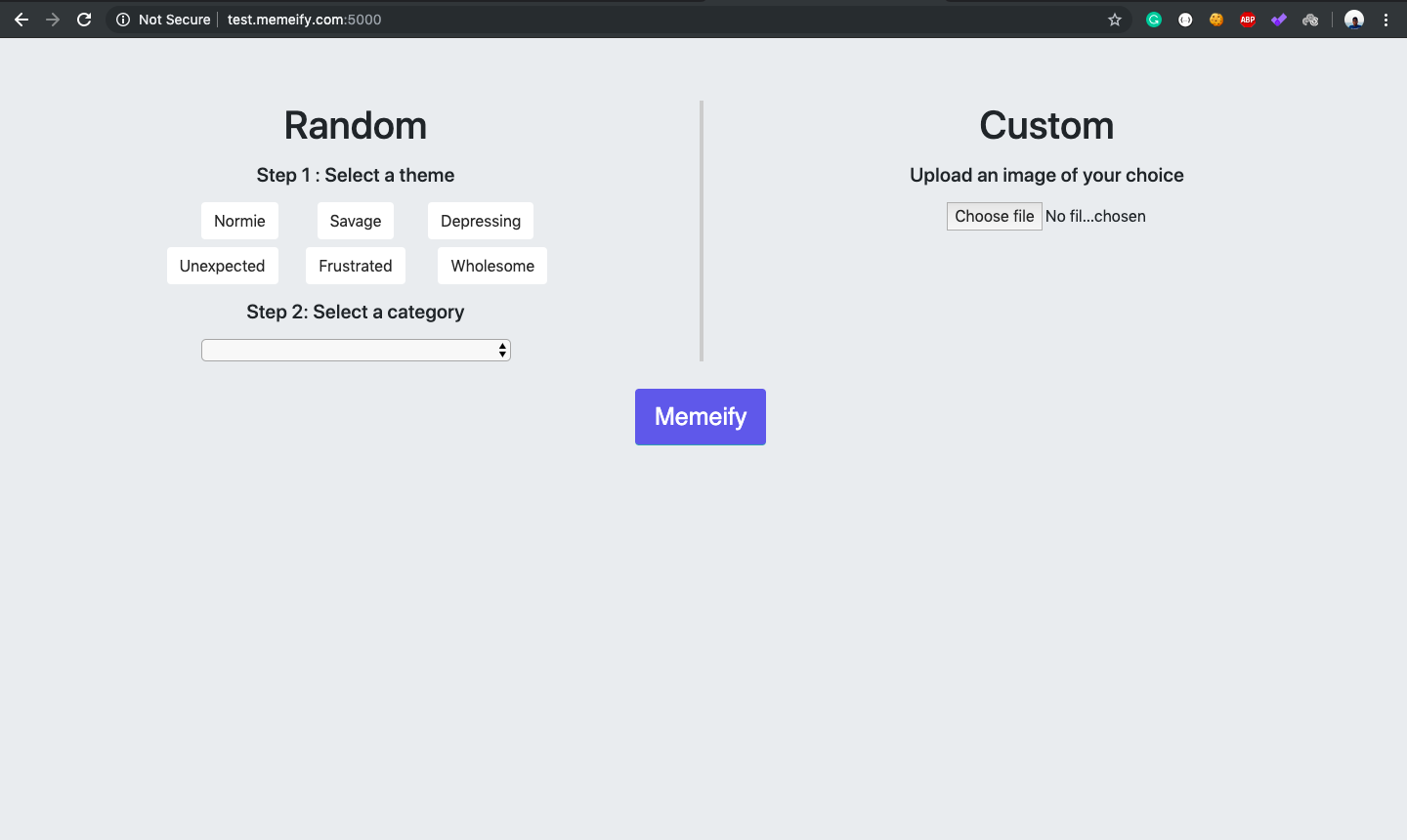}}\quad
  \subfigure[Memeify web application showing a meme]{\includegraphics[width=.40\linewidth,height=.30\linewidth]{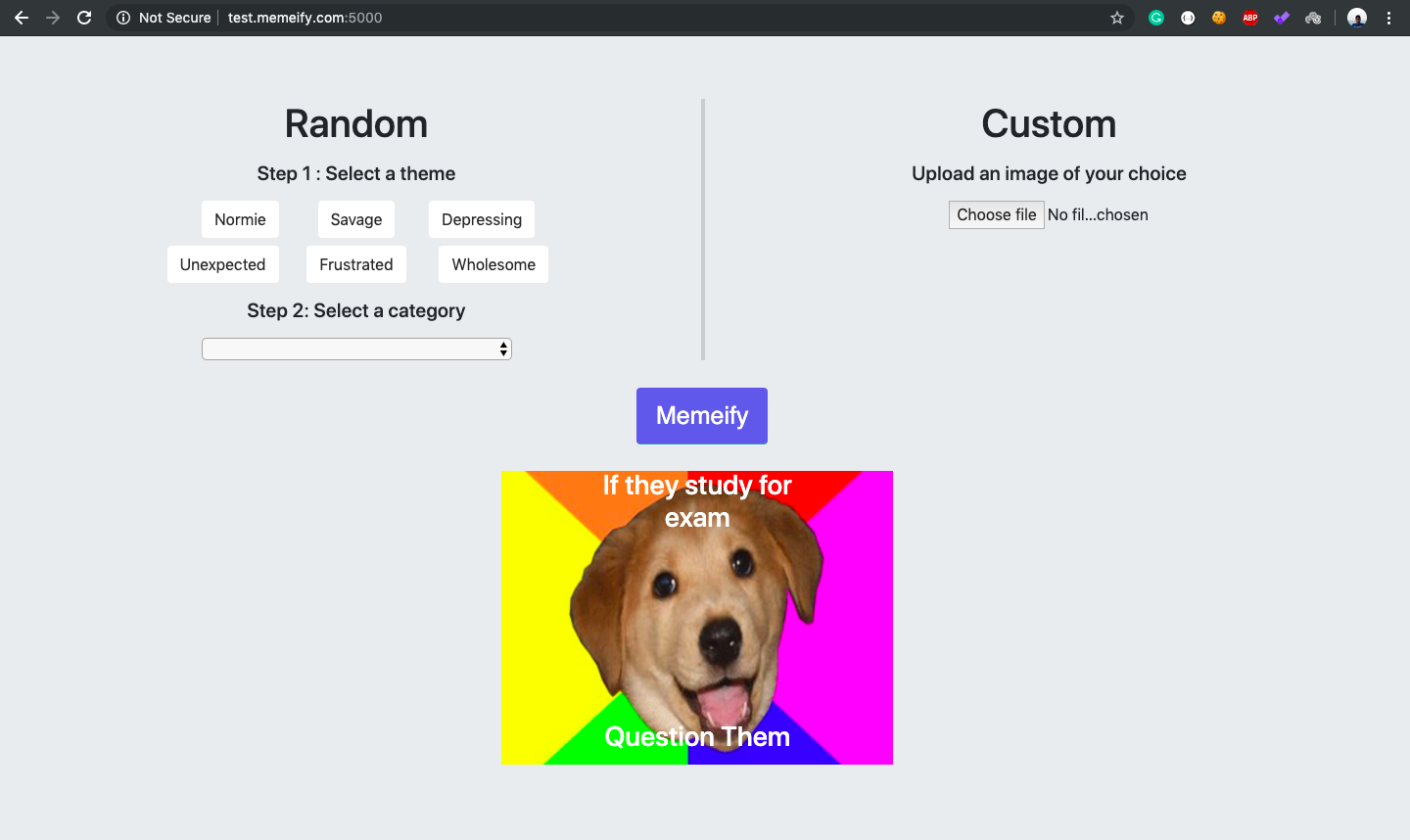}}
  \caption{Memeify -- demo system images.}
  \label{fig:demo}
\end{figure*}

\subsection{Generation Model} \label{ssec:genm}
We consider the problem of meme generation as a language modeling task to produce funny and apt captions when given an input image or a class label as a prompt. We train multiple deep learning models, i.e., LSTM networks \cite{Hochreiter:1997:LSM:1246443.1246450} and their variants, for modeling the text from our meme corpus. However, qualitative analysis of the generated captions reveal that most of the models have the following limitations:
\begin{itemize}
    \item They do not accurately capture the class information.
    \item They are not able to reproduce humour well.
\end{itemize}

To mitigate these issues, we use the transformer based GPT-2 architecture \cite{radford2019language} as our base language generative model. 
We incorporate the class information for the different memes by prepending a particular meme caption with its class name. This helps us generate class specific meme captions by enforcing the model to use the class information. The GPT-2 architecture also solves the issue of expressing humour in the text due to its self-attention capabilities \cite{DBLP:journals/corr/VaswaniSPUJGKP17}, large-scale generative pre-training and multiple-task adaptability. 

We use this generative model trained with the class information as the caption generator in the Memeify system. A few examples of the generated captions are shown in Table \ref{tab: examples_data}.

The Memeify system enables the generation of memes in two specific ways:
\begin{enumerate}
    \item {\bf Randomization:} As explained in Section \ref{dataset}, every meme that we generate has an associated class and theme. We enable the users to randomly pick a theme and a class, and we then use this class as a seed to the generative model which then produces an appropriate caption. The corresponding image for the meme is retrieved from a database of default meme class images.
    \item {\bf Customization:} A user can upload custom images into the system, and the system can aptly produce captions fitting the image. To correctly identify the theme and class of the required meme, we use a similarity matching algorithm that classifies the image into its correct contextual theme. A pretrained VGG16 convnet \cite{simonyan2014very} is used as a feature extractor to convert images to feature vectors. We extract feature vectors for each of the default class images from the database and then use a locality sensitive hashing algorithm to create a lookup table. For every new image uploaded into the system, we convert it into a feature vector and then obtain the meme class by referring to the lookup table. The obtained class is then used as a seed to the generative model for the caption generation. 
\end{enumerate}

\begin{figure}
  \includegraphics[width=\linewidth]{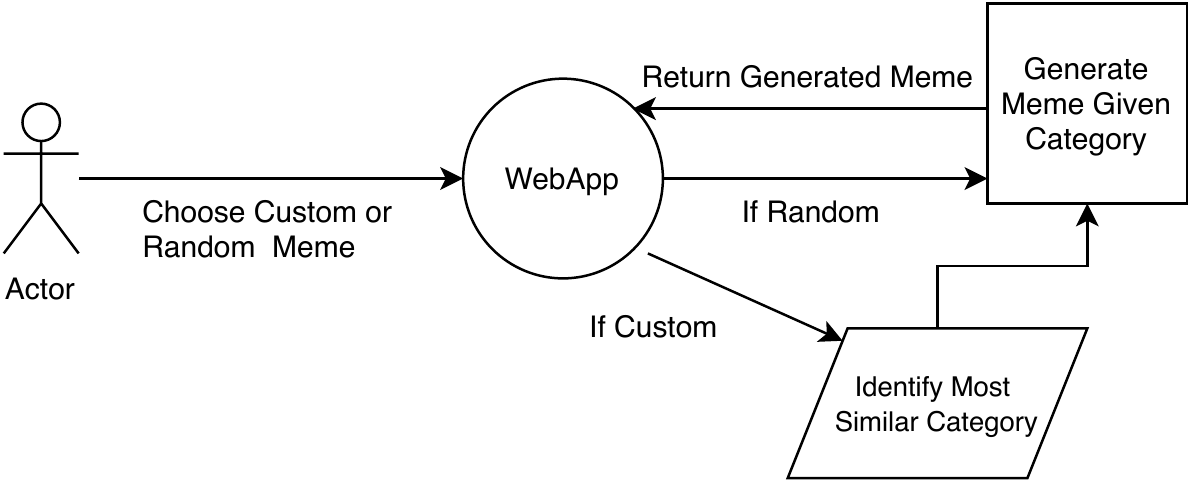}
  \caption{Architecture of Memeify.}
  \label{fig:boat1}
  \vspace{-5mm}
\end{figure}

\subsection{Web Application}
We develop a web interface for users to interact with our meme generation algorithm. The landing page of the website is shown in Figure \ref{fig:demo}. The users are provided with two options based on the model explained above:
\begin{enumerate}
    \item {\bf Random Image:} The users can choose to generate a random meme by selecting a theme and a class on the website. This data is sent to the backend written in Flask that uses the trained model to generate memes.
    \item {\bf Custom Image:} The users can also generate a meme caption for an image of their choice. The user uploads an image   onto the website. The image is sent to our server, and a caption is generated.
\end{enumerate}
\subsection{Engineering}
\begin{enumerate}
    \item The generation model is the bottleneck in the pipeline. Therefore, we implement a redis cache on our backend for the generated memes. We ensure that memes are not repeated for a single user by using web sessions.
    \item All data transfers to and from the user's webpage to the server happen asynchronously using AJAX. This ensures a seamless experience for the user with minimum delay.
\end{enumerate}
The complete pipeline is shown in Figure \ref{fig:boat1}.

\section{Evaluation and Analysis}

\begin{table*}[t]
\centering
\caption{Examples from the dataset and generation system showing default image, original captions, generated captions using Memeify, class and theme.} 
\begin{tabular}{  |m{4cm} | m{4cm} | m{4cm} | m{1.7cm} | m{1.7cm}| }
\hline
\multicolumn{1}{|r|}{\textbf{Default image}} & \multicolumn{1}{c|}{\textbf{Original caption}} & \textbf{Memeify-generated caption} & \multicolumn{1}{c|}{\textbf{Class}} & \multicolumn{1}{c|}{\textbf{Theme}} \\ \hline
\begin{minipage}{.3\textwidth}
    \centering
      \includegraphics[width=15mm, height=15mm]
      {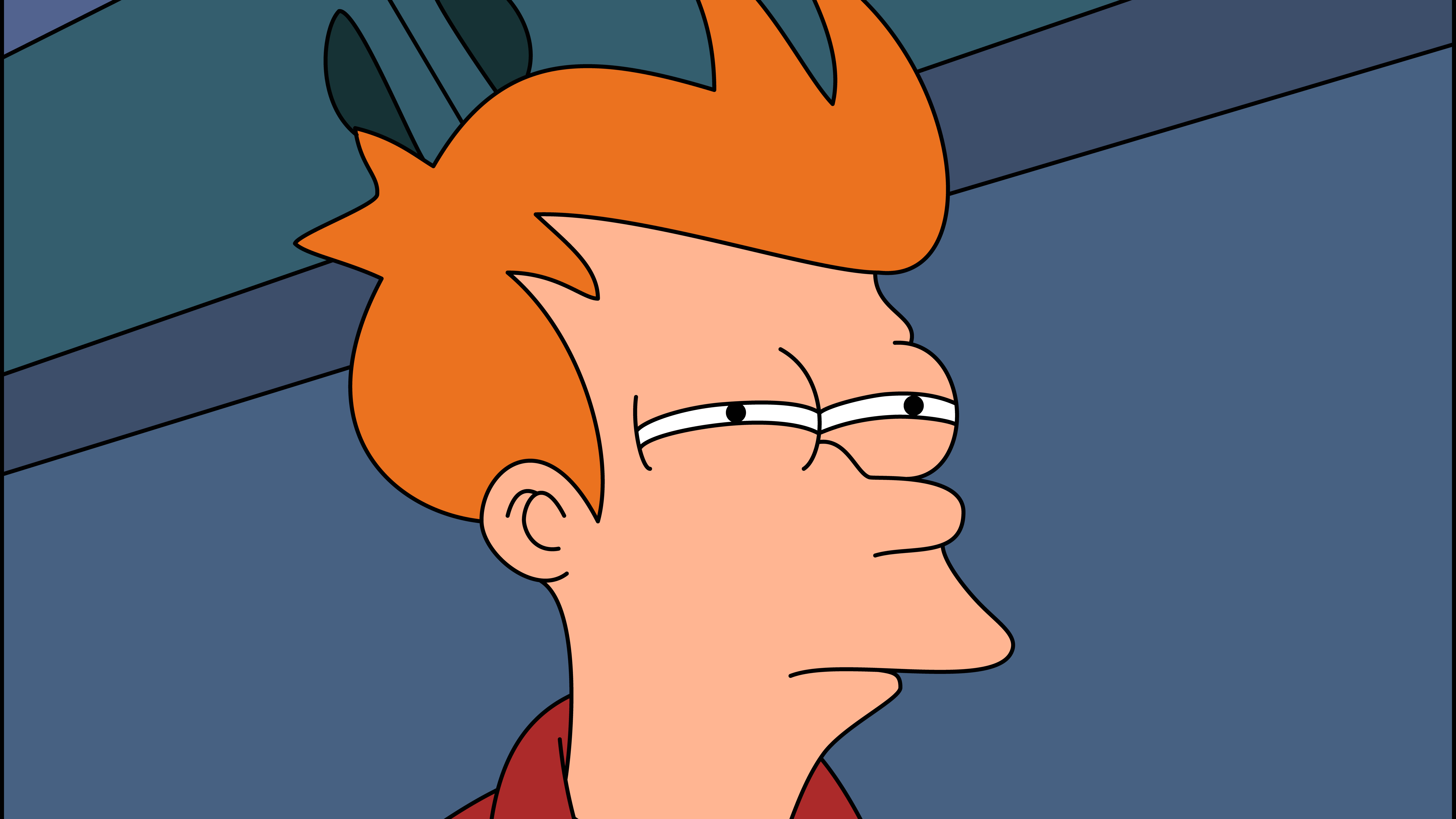}
    \end{minipage}         & 
    \begin{itemize}
        \item not sure if smart or just british
        \item not sure if joking or serious
      \end{itemize} &
      \begin{itemize}
        \item not sure if trolling or stupid
        \item not sure if intelligent or surrounded by idiots 
      \end{itemize} &
      futuruma\_fry &
      Frustrated
    \\ \hline
\begin{minipage}{.3\textwidth}
    \centering
      \includegraphics[width=15mm, height=15mm]{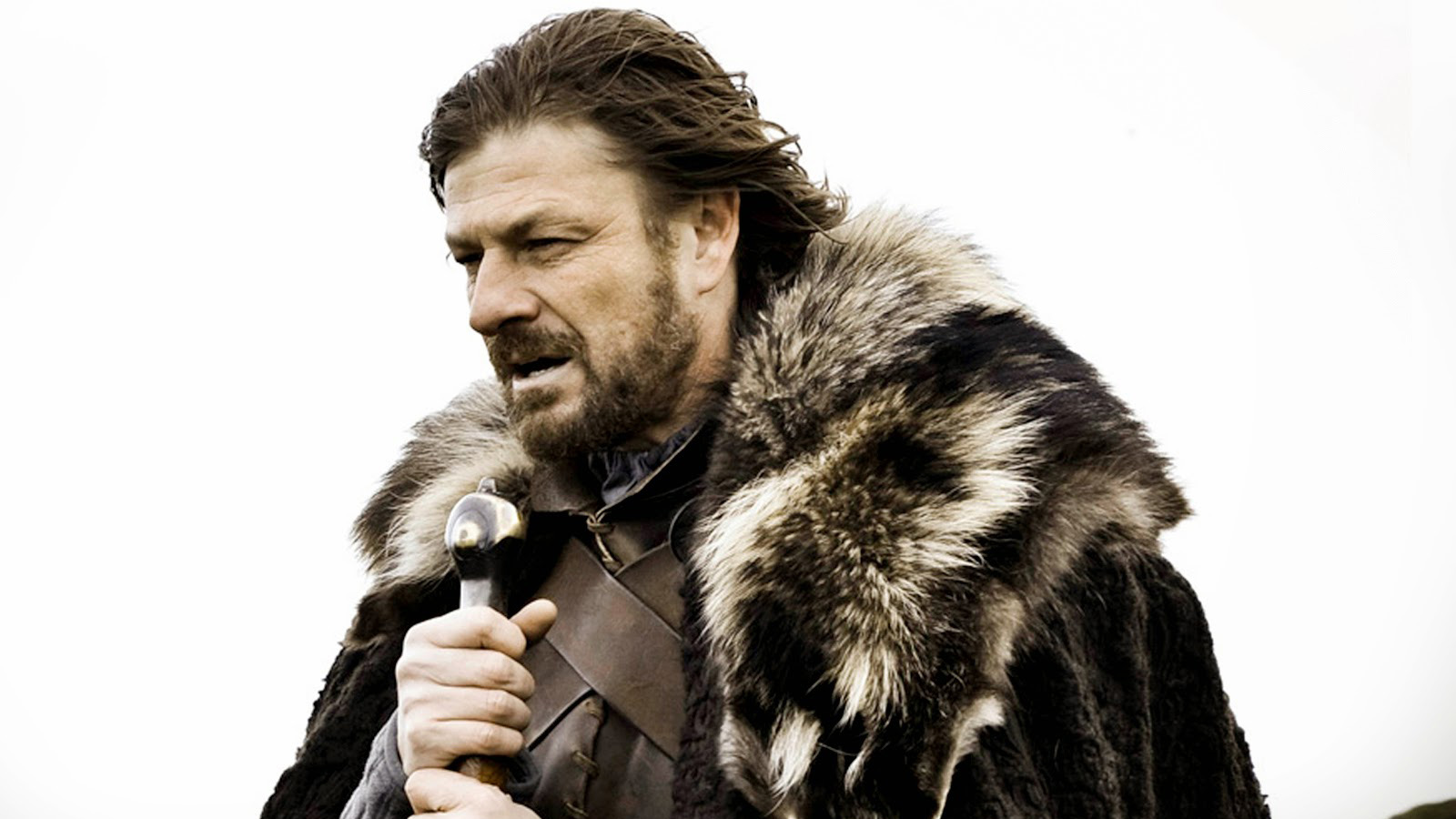}
    \end{minipage}         & 
    \begin{itemize}
            \item brace yourselves winter is coming
            \item brace yourselves shitty memes are coming
         \end{itemize} &
        \begin{itemize}
            \item brace yourselves 9 AM snoozes are coming
            \item brace yourselves email overload is coming
      \end{itemize} &
      imminent\_ned &
        Normie           \\ \hline
\end{tabular}
\label{tab: examples_data}
\end{table*}

\label{analysis}
To effectively evaluate the quality of the memes generated and the robustness of the generative model, we performed a human-evaluation by conducting a user study. 20 human experts\footnote{They were social media experts, and their age ranged from 25 to 40 years.} volunteered for the study. We performed three different analysis/evaluation tasks to qualitatively understand the working of the generative model. We consider the Dank Learning meme generator proposed by Peirson et al. \cite{peirson2018dank} as a baseline model for a comparative evaluation.
The evaluative tasks are explained in detail in the remaining part of the section.

\subsection{User Satisfaction}
\label{user-satisfaction}
To evaluate user satisfaction levels, we conducted a rating study.
We generated a batch of 100 memes for each theme by randomly picking classes within each theme, individually for the baseline model and our model. For each theme, we showed a set of 5 generated memes from our model, 5 generated memes from the baseline model, and 5 original memes, in a mixed order (without revealing the identity of the original and generated memes), to each volunteer. We asked them to rate the generated and original memes from a range of $1$ (lowest rating) to $5$ (highest rating) based on caption-content, humour and originality. For each theme, we then averaged out the results from the 20 volunteers as shown in Table \ref{tab: user_sat}. We observe that the quality of our generated memes is almost on par with the original memes. We also notice that our model outperforms the baseline model, across all themes. 
Therefore, we maintain this on a holistic level; all of the volunteers were satisfied with the quality of the generated memes as well.

\begin{table}[!h]
\caption{Average ratings of volunteers for original and generated memes (baseline and ours) across themes. ARO represents average ratings for original memes, ARG represents average ratings for generated memes (our model), and ARB represents average ratings for generated memes (baseline model).}
\vspace{-3mm}
\begin{tabular}{|l|c|c|c|}
\hline
\textbf{Theme}      & \textbf{ARO} & \textbf{ARG} & \textbf{ARB}\\ \hline
Normie     & 3.1                              & 2.9                 & 2.7              \\ \hline
Savage     & 3.6                              & 3.6                 & 3.4              \\ \hline
Depressing & 3.2                              & 3.1                 & 3              \\ \hline
Unexpected & 3.5                              & 3.3                 & 3.3              \\
\hline
Frustrated & 3.2                              & 3                   & 2.9              \\
\hline
Wholesome  & 2.8                              & 2.7                 & 2.6              \\
\hline
Overall    & 3.23                             & 3.1           & 2.98\\ \hline 
\end{tabular}
\label{tab: user_sat}
\vspace{-5mm}
\end{table}


\subsection{Differentiation Ability of Users}
\label{differention}
\begin{figure}
  \subfigure[our model]{\includegraphics[scale=0.3]{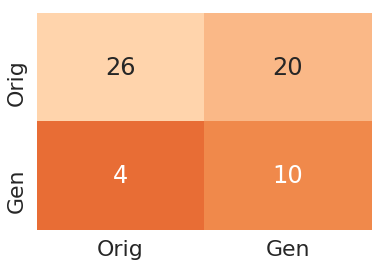}}\quad
  \subfigure[baseline model]{\includegraphics[scale=0.3]{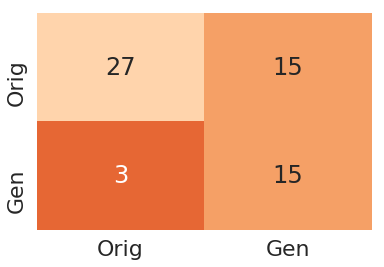}}
  \vspace{-5mm}
  \caption{Confusion matrices for the baseline model and our model.}
  \label{fig:confusion_mat}
\end{figure}

We wanted to understand if the memes generated by our model were significantly different from the ones in the dataset. For this purpose, we generated a separate batch of 50 memes for each theme, randomly picking a class within each theme, for the baseline model and our model (similar to the generation process in Section \ref{user-satisfaction}). For each theme, we showed 5 generated memes and 5 original memes to volunteers and asked them to classify a meme as generated or original. We then analysed the results of this classification study, plotted confusion matrices for both the baseline model and our model as shown in Figure \ref{fig:confusion_mat} and reported the results of the individual evaluation metrics as shown in Table \ref{tab: metrics}.

From the confusion matrix of our model, we observe that the volunteers mis-classified generated memes as original memes $66.67\%$ of the time. This vouches for the authenticity of our model and shows that the generated memes are qualitatively humorous and genuine enough to confuse the volunteers between original and generated memes. 

\begin{table}[h]
\caption{Evaluation (based on precision, recall, accuracy and F1-score)  of the classification study presented in Section \ref{differention}.}
\vspace{-3mm}
\begin{tabular}{|l|l|l|}
\hline
\textbf{Metric}    & \textbf{Baseline} & \textbf{Our Model} \\ \hline
Precision & 64.28   & 56.52 \\ \hline
Recall   & 90 & 86.66 \\ \hline
Accuracy & 70 & 60    \\ \hline
F1-score & 75.0 & 68.42 \\ \hline
\end{tabular}
\label{tab: metrics}
\vspace{-5mm}
\end{table}
    
\subsection{Theme Recovery}
\label{theme-recovery}
We also wanted to understand if the idea of themes as `broad categories' (explained in Section \ref{theme_data}) held true across volunteers. To understand this, we hypothesised that if volunteers were correctly able to recover themes from randomly sampled memes, then it would empirically prove the reasoning behind the formation of themes. To acquaint volunteers with the idea of a theme, we showed a set of 10 memes corresponding to each theme from our dataset.

We then generated a batch of $100$ memes (from our model) across themes, randomly picking classes (similar to generation processes in Sections \ref{user-satisfaction} and \ref{differention}). We asked the volunteers to classify a sample of 20 generated memes into the 6 themes. We report the overall classification accuracy and per-theme classification accuracy averaged out across all the volunteers in Table \ref{tab: theme_recovery}.
We notice the high overall classification accuracy which confirms our hypothesis about the existence of themes across meme classes. On further examination, we notice that individually all the themes have very high accuracies apart from the `Normie' theme. We attribute the relatively low accuracy of the `Normie' theme, to the large number of classes (refer Table \ref{tab: theme_recovery}) having low intra-theme similarity as compared to other themes ({\em cf.} Figure \ref{fig:cluster} (a)). 
Hence, our hypothesis of existence of themes across meme classes is validated.


\begin{table}[h]
\caption{Overall accuracy and per-theme accuracy for the classification study of Section \ref{theme-recovery}.}
\begin{tabular}{|l|l|}
\hline
\textbf{Theme}      & \textbf{Accuracy} \\ \hline
Normie     & 77.3     \\ \hline
Savage     & 86.1     \\ \hline
Depressing & 84.6     \\ \hline
Unexpected & 90.2     \\ \hline
Frustrated & 87.7     \\ \hline
Wholesome  & 86.8     \\ \hline
Overall    & 85.5 \\ \hline
\end{tabular}
\label{tab: theme_recovery}
\vspace{-5mm}
\end{table}

\section{Conclusion}
In this paper, we explained our meme generation system - Memeify. Memeify is capable of generating memes either from existing classes and themes or from custom images. We also created a large-scale meme dataset consisting of meme captions, classes and themes. We provided an in-depth qualitative analysis on the basis of a user-study. Currently, our work is only limited to memes which have up to two parts in the meme caption. However, we are interested in extending the Memeify system to include multiple parts in the meme caption.


\bibliographystyle{ACM-Reference-Format}
\bibliography{sample-base}


\end{document}